\documentclass[10pt,twocolumn,letterpaper]{article}

\usepackage{cvpr}
\usepackage{times}
\usepackage{epsfig}
\usepackage{mathtools}
\usepackage{graphicx}
\usepackage{amsmath}
\usepackage{amssymb}
\usepackage[table,xcdraw]{xcolor}
\usepackage{multirow}
\usepackage[breaklinks=true,bookmarks=false]{hyperref}

\cvprfinalcopy % *** Uncomment this line for the final submission

\def\cvprPaperID{****} % *** Enter the CVPR Paper ID here

\ifcvprfinal\fi
\begin{document}

%%%%%%%%% TITLE
\title{SREdgeNet: Edge Enhanced Single Image Super Resolution\\ using Dense Edge Detection Network and Feature Merge Network}

\author{Kwanyoung Kim, \space \space \space Se Young Chun\\
Ulsan National Institute of Science and Technology (UNIST), Republic of Korea\\
{\tt\small \{cubeyoung, sychun\}@unist.ac.kr}}

\maketitle
\thispagestyle{empty}

%%%%%%%%% ABSTRACT
\begin{abstract}
Deep learning based single image super-resolution (SR) methods have been rapidly evolved over the past few years and have yielded state-of-the-art performances over conventional methods. Since these methods usually minimized $l1$ loss between the output SR image and the ground truth image, they yielded very high peak signal-to-noise ratio (PSNR) that is inversely proportional to these losses. Unfortunately, minimizing these losses inevitably lead to blurred edges due to averaging of plausible solutions. Recently, SRGAN was proposed to avoid this average effect by minimizing perceptual losses instead of $l1$ loss and it yielded perceptually better SR images (or images with sharp edges) at the price of lowering PSNR. In this paper, we propose SREdgeNet, edge enhanced single image SR network, that was inspired by conventional SR theories so that average effect could be avoided not by changing the loss, but  by changing the SR network property with the same $l1$ loss. Our SREdgeNet consists of 3 sequential deep neural network modules: the first module is any state-of-the-art SR network and we selected a variant of EDSR~\cite{lim2017enhanced}. The second module is any edge detection network taking the output of the first SR module as an input and we propose DenseEdgeNet for this module. Lastly, the third module is merging the outputs of the first and second modules to yield edge enhanced SR image and we propose MergeNet for this module.
Qualitatively, our proposed method yielded images with sharp edges compared to other state-of-the-art SR methods. Quantitatively, our SREdgeNet yielded state-of-the-art performance in terms of structural similarity (SSIM) while maintained comparable PSNR for $\times$8 enlargement.
\end{abstract}

%%%%%%%%% BODY TEXT
\section{Introduction}

\begin{figure}[t]
    \begin{center}
    \centering
	\includegraphics[width=1\linewidth]{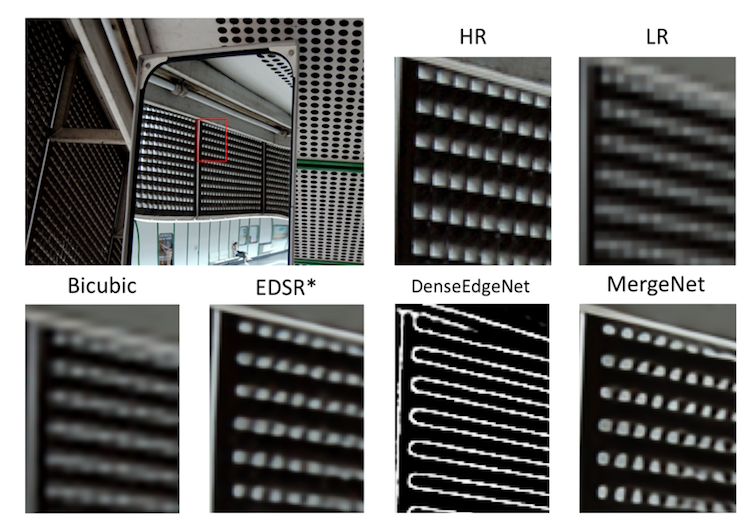}
	\caption{\protect Illustration of SREdgeNet pipeline. Firstly, a upscale SR image is produced from a LR image by EDSR*. Then, an edge image is obtained from the SR image by our edge detection network, DenseEdgeNet. 	
Finally, our MergeNet that combines two previous outputs yields the final SR image with edge enhancement.}
	\label{fig:Figure1}
    \end{center}	
\end{figure}

The goal of single image super-resolution (SR) is to recover a high resolution (HR) image from a low resolution (LR) image.
Single image SR problem from a LR image is challenging and ill-posed problem since lost high frequency information in the LR image must be restored. 
Large enlargement task in SR (e.g., $\times$8 enlargement) is especially challenging due to massive amount of lost high frequency information compared to other small enlargement tasks (e.g., $\times$2-4 enlargements).
To solve the SR problem, deep learning based methods have recently been proposed. They significantly improved performances over conventional SR methods in terms of peak signal-to-noise ratio (PSNR). 
There are several works on single image SR deep neural networks such as %, for example, 
SRCNN~\cite{dong2016image}, VDSR~\cite{kim2016accurate}, SRResNet~\cite{ledig2017photo}, EDSR~\cite{lim2017enhanced}, and D-DBPN~\cite{haris2018deep}. D-DBPN achieved 
state-of-the-art performance in $\times$8 enlargement.

Although these methods quantitatively improved performances in PSNR by minimizing the mean squared error (MSE) between ground truth and SR output, minimizing MSE or $l1$ loss tends to yield blurred edges due to the pixel-wise averages of all plausible solutions~\cite{ledig2017photo}. This averaging effect often resulted in losing much HR components, especially detailed edges. Recently proposed SRGAN showed that using perceptual losses yielded qualitatively (perceptually) better and more realistic SR images with sharp edges while it yielded substantially lower performance in PSNR compared to other state-of-the-art SR methods.

It is desirable to use $l1$ or MSE losses for quantitatively high performance since these losses are inversely proportional to quantitative measures such as PSNR. At the same time, it is also desirable to avoid average effect by yielding one plausible solution instead of averaging all possible solutions. Inspired by a theoretical SR approach~\cite{Ongie:2016jqa}, 
we proposed SREdgeNet, a novel edge enhanced single image SR network that can take advantage of both quantitatively high performance using $l1$ loss and qualitatively high performance by providing edge information as a guide to select less number of plausible solutions using edge enhanced networks.
Figure~\ref{fig:Figure1} describes the pipeline of our proposed method.
Firstly, a upscale SR image is produced from a LR image by EDSR*, a variant of EDSR~\cite{lim2017enhanced}. Then, an edge image is obtained from the SR image by our novel edge detection network, DenseEdgeNet. 	
Finally, our MergeNet that combines two previous outputs yields the final SR image
by reinforcing edge information.
Our method should be more effective for large factor enlargement (e.g., $\times 8$) that has much more plausible solutions at each pixel. 
 In summary, our main contributions are three-fold:
 
 \begin{figure}[t]
    \begin{center}
    \centering
	\includegraphics[width=1\linewidth]{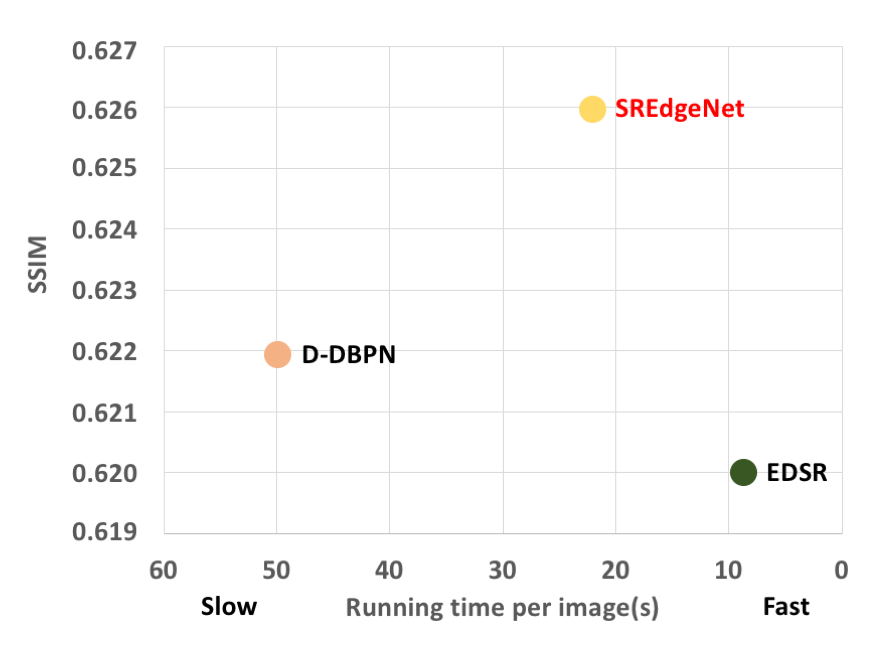}
	\caption{\protect SSIM vs Running time per image. Comparison with other networks.~\cite{lim2017enhanced,haris2018deep} The results are evaluated with Urban100 dataset for $\times$8 enlargement.}
	\label{fig:Figure2}
    \end{center}	
\end{figure}

1) We propose SREdgeNet, an edge enhanced single image SR network, consisting of three modules. In the first module, a variant of EDSR~\cite{lim2017enhanced} (EDSR*) was developed by incorporating pyramid pooling into the upsampling layer of EDSR for better exploiting multiscale features.
Our proposed SREdgeNet worked well especially for $\times 8$ enlargement in structural similarity (SSIM) with relatively fast computation time
as illustrated in Figure~\ref{fig:Figure2}.

2) We propose DenseEdgeNet, a novel edge detection network that is composed of dense residual blocks, has several dense connections to take advantage of global feature map information, and uses a multi-complexity method to obtain rich feature map information.

3) We propose MergeNet that exploits the outputs of EDSR* and DenseEdgeNet to yield the final SR image. Shallow and deep features are jointly combined with residual learning and edge skip connection was introduced to allow edges to be well-fused with SR image.
%-------------------------------------------------------------------------

%-------------------------------------------------------------------------
\section{Related Works}
\subsection{Deep learning based super resolution}

Recently, deep learning based methods have achieved significant improvements in SR problems against other conventional non-deep learning based methods. Dong \textit{et al.} proposed SRCNN~\cite{dong2016image} that established an end-to-end manner mapping between a upscaled LR image using bicubic interpolation and its HR counterpart. Kim \textit{et al.} proposed VDSR~\cite{kim2016accurate}, a deep neural network using residual learning, that yielded improved PSNR performance over SRCNN. The network of this method was trained to yield a residual image that is the difference between an interpolated LR image and a ground truth HR image.
DRCN~\cite{kim2016deeply} was proposed with very deep recursive layers with recursive learning and DRRN~\cite{tai2017image} was also introduced with a recursive block with 52 convolutional layers. All of these methods use an interpolated or upsampled LR image as the input to the networks. 
Lai \textit{et al.} proposed a Laplacian pyramid SR network (LapSRN)~\cite{lai2017deep} that progressively reconstructed the multiple images with different scales, for example, starting from $\times$8 to $\times$4 to $\times$2.

Legit \textit{et al.} proposed SRResNet~\cite{ledig2017photo} using residual blocks to significantly increase the size of receptive field and to include local context information. SRGAN~\cite{ledig2017photo} was also proposed with the same network structure as SRResNet, but with a different training loss function based on a discriminator network for perceptual loss. SRGAN yielded visually (perceptually) pleasing SR outputs while PSNR of SRGAN was substantially lower than that of SRResNet. SRResNet yielded an average of many plausible SR outputs while SRGAN yielded one of many possible SR outputs.

Lim \textit{et al.} proposed EDSR that enhanced SRResNet by eliminating batch normalization and by stacking deeper layers (residual blocks from 16 to 32, filter channels from 64 to 256)~\cite{lim2017enhanced}. EDSR was trained with used $l1$ loss instead of $l2$ loss for better PSNR. This network produced good results in terms of PSNR, but lost much high frequency information due to averaging effect of many plausible solution so that tiny texts and edges are blurred out. %To resolve this issue, 
Haris \textit{et al.} proposed D-DBPN~\cite{haris2018deep} that introduced iterative up and down sampling layers rather than emphasizing the existing feature extraction by stacking deeper convolutional layers. By providing a feedback for error projection, D-DBPN achieved state-of-the-art performance.

\subsection{Edge detection}

Edge Detection is one of the most fundamental problems in computer vision.
Early conventional methods focused primarily on color intensity and gradients. The Sobel operator was proposed~\cite{sobeledge} that computed the gradient map of an image and then produced an edge map by thresholding. Further, Canny~\cite{canny1986computational} introduced Gaussian smoothing in the process of extracting the gradient of image and used the bi-thresholding to get a more robust edge map. This method is still efficient and is used to obtain detailed edges. However, these early methods are less accurate and less suitable for diverse modern applications.

Recently, deep learning based methods have been proposed to solve edge detection problems and achieved significant performance improvements.  Related works include deep contour~\cite{shen2015deepcontour}, deep edge~\cite{bertasius2015deepedge}, $N^4$ fields~\cite{ganin2014n} and CSCNN\cite{hwang2015pixel}, and HED~\cite{xie2015holistically}. Among them, the HED method predicted an edge map from an image and % to image fashion and
established an end-to-end manner for training. Thereafter, many deep learning based edge detection methods have been proposed based on HED~\cite{liu2017richer}.
Aforementioned deep learning based methods have improved edge detection performance, %the best methods, 
but have lost much feature map information in the process of predicting edges. To resolve this issue, we propose 
DenseEdgeNet, a novel edge detection network that has a dense connection to prevent information lost in the hierarchical CNN features and to yield more precise edge information by introducing a multi-complexity method.

\subsection{Connections with high level vision tasks}

Joint estimation method has been used to improve performance for image processing and computer vision. Especially, linking between low level vision task and high level vision task often resulted in better performances. Recently, Liu \textit{et al.} proposed a method connecting image denoising and image segmentation based on deep learning~\cite{liu2018connecting}. 
To enhance both the low level vision task, which is image denoising, and relatively high level vision task, which is image segmentation, they cascaded two modules for jointly performing two tasks to synergetically improve both performances.

Similar to it, utilizing high level vision task have been proposed in SR. Makwana \textit{et al.}~\cite{makwana2013single} proposed a non-deep learning method that introduces iterative feedback structure using Canny edge detector~\cite{canny1986computational}. 
DEGREE~\cite{yang2017deep} that progressively reconstructed SR images with the help of edge-guiding was also proposed. This method firstly produced an edge map from a LR image by using Sobel operator~\cite{sobeledge} and then combined this edge image with an interpolated image by using recurrent neural network. Ongie \textit{et al.} proposed a theoretically well-grounded SR method that firstly estimated edges using level set method and then reconstructed a SR image using edge information~\cite{Ongie:2016jqa}.
Inspired by these methods, we propose to connect SR with edge detection.
% corresponding to high vision task.
%------------------------------------------------------------------------

%-------------------------------------------------------------------------
\begin{figure*}[t]
    \begin{center}
    \centering
	\includegraphics[width=0.95\linewidth]{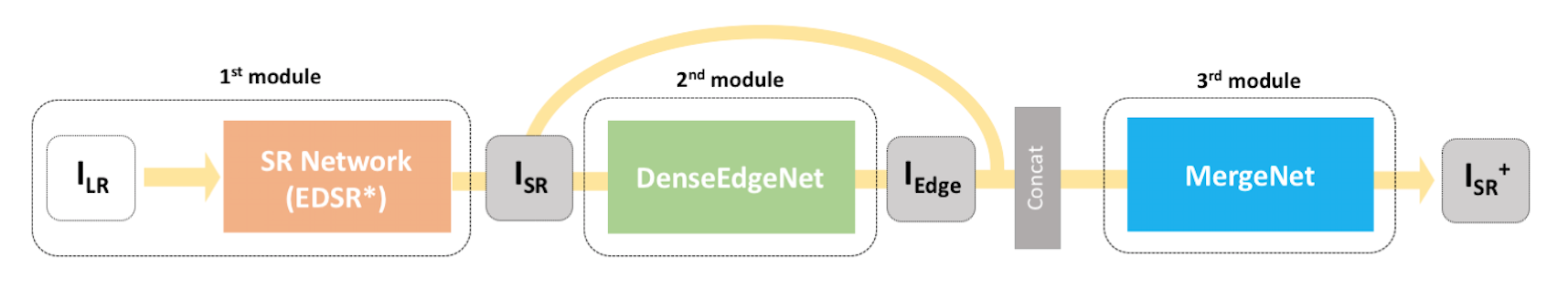}
	\caption{\protect Our proposed SREdgeNet, consisting of 3 modules for single image SR including EDSR*, DenseEdgeNet, and MergeNet.} 
	\label{fig:overall}
    \end{center}	
\end{figure*}

%-------------------------------------------------------------------------
\begin{table*}[!t]
\begin{center}
\begin{tabular}{|c|c|c|c|c|c|l|l|l|l|}
\hline
          &       & \multicolumn{2}{c|}{Set5}                                   & \multicolumn{2}{c|}{Set14} & \multicolumn{2}{c|}{BSD100}                                 & \multicolumn{2}{c|}{Urban100}                               \\ \hline
Algorithm & scale & PSNR                         & SSIM                         & PSNR         & SSIM        & PSNR                         & SSIM                         & PSNR                         & SSIM                         \\ \hline
EDSR      & 4     & 32.46                        & 0.897                        & {\color[HTML]{FE0000}28.80}        & {\color[HTML]{FE0000}0.788}       & 27.61                        & 0.737                        & 26.25                        & 0.795                        \\ \hline
EDSR*   & 4     & {\color[HTML]{FE0000} 32.57} & {\color[HTML]{FE0000} 0.898} & 28.40        & 0.780       & {\color[HTML]{FE0000} 27.62} & {\color[HTML]{FE0000} 0.737} & {\color[HTML]{FE0000} 26.39} & {\color[HTML]{FE0000} 0.790} \\ \hline
EDSR      & 8     & 26.97                        & 0.775                        & {\color[HTML]{FE0000}24.91}        & {\color[HTML]{FE0000}0.640}       & 24.80                        & 0.596                        & 22.47                        & 0.620                        \\ \hline
EDSR*   & 8     & {\color[HTML]{FE0000} 27.12} & {\color[HTML]{FE0000} 0.782} & 24.80        & 0.640       & {\color[HTML]{FE0000} 24.84} & {\color[HTML]{FE0000} 0.599} & {\color[HTML]{FE0000} 22.56} & {\color[HTML]{FE0000} 0.622} \\ \hline
\end{tabular}
\end{center}
\caption{\protect Comparison of EDSR* with EDSR~\cite{lim2017enhanced} on benchmark datasets. {\color[HTML]{FE0000}Red} indicates better results than the other.} 
\label{table1}
\end{table*}

\section{Methods}
In this section, we describe our proposed network for SR. Let us denote 
$I_{LR}$ and $I_{SR}$ as the input and output images of a SR Network.  
Our proposed SREdgeNet network is divided into three modules as illustrated in Figure~\ref{fig:overall}.

The first module, SR network, takes %In the first module, as a SR Network, it takes  
$I_{LR}$ as an input and upscales it to the desired size as follows:
\begin{center}
 $I_{SR}$ = SR($I_{LR}$)
\end{center}
where SR($\cdot$) denotes the function of deep SR network. We implemented this first SR Network module by modifying EDSR~\cite{lim2017enhanced} to use pyramid pooling that integrates both local and global information. We will denote this SR Network as EDSR*, a variant of EDSR.
%used EDSR* to implement SR Network. EDSR* is enhanced version of  by using 

The second module is an edge detection network. We propose DenseEdgeNet, an edge detection network that predicts edges from the upscaled SR image, which is the output of the first SR module:
\begin{center}
$I_{Edge}$ = E($I_{SR}$)
\end{center}
where $I_{Edge}$ denotes the output of DenseEdgeNet and E($\cdot$) denotes the function of edge detection network.

The final module integrates $I_{SR}$ and $I_{Edge}$ in our MergeNet. Our proposed MergeNet concatenates $I_{SR}$ and $I_{Edge}$, and then produces the final SR image, $I_{SR}^+$ as:
\begin{center}
$I_{SR}^+$ = M($I_{SR}$$\bigodot$$I_{Edge}$)
\end{center}
where $\bigodot$ denotes concatenation operation and M($\cdot$) denotes the function of MergeNet with %that combine 
both $I_{SR}$ and $I_{Edge}$.

\subsection{EDSR*}
%We introduce SRNetwork, EDSR-PP\cite{park2018efficient}, that is  based on EDSR~\cite{lim2017enhanced}. 
EDSR* incorporates pyramid pooling in the up-sampling layer of EDSR~\cite{lim2017enhanced}. In general, the receptive field size of deep learning-based image processing implies how much feature map information is contained and used. More stacks of convolutional layers are typically leading to larger receptive field sizes. However, among existing deep learning based networks for SR that stack deeper layers such as EDSR~\cite{lim2017enhanced} or SRResNet~\cite{ledig2017photo}, there is no method to utilize global context information.

To resolve this issue, pyramid polling~\cite{zhang2018residual} method was proposed for utilizing both local and global context information. We modified the original EDSR by incorporating pyramid polling into the SR network. We denote this variant of EDSR as EDSR*.
%We incorporate it into EDSR to address this SR issue. 
Unlike the upsampling layer in EDSR, EDSR* first performs average pooling and executes convolutions for each of the four pyramid scales. Then these are stacked in the previous feature map.  We used %replaced 4x4 with 
6$\times$6 pyramid pulling to receive more enlarged information. Comparing with EDSR, our modified EDSR* yielded better performance than EDSR as shown in Table~\ref{table1}. 
%-------------------------------------------------------------------------

%-------------------------------------------------------------------------
\begin{figure*}[t]
    \begin{center}
    \centering
	\includegraphics[width=0.7\linewidth]{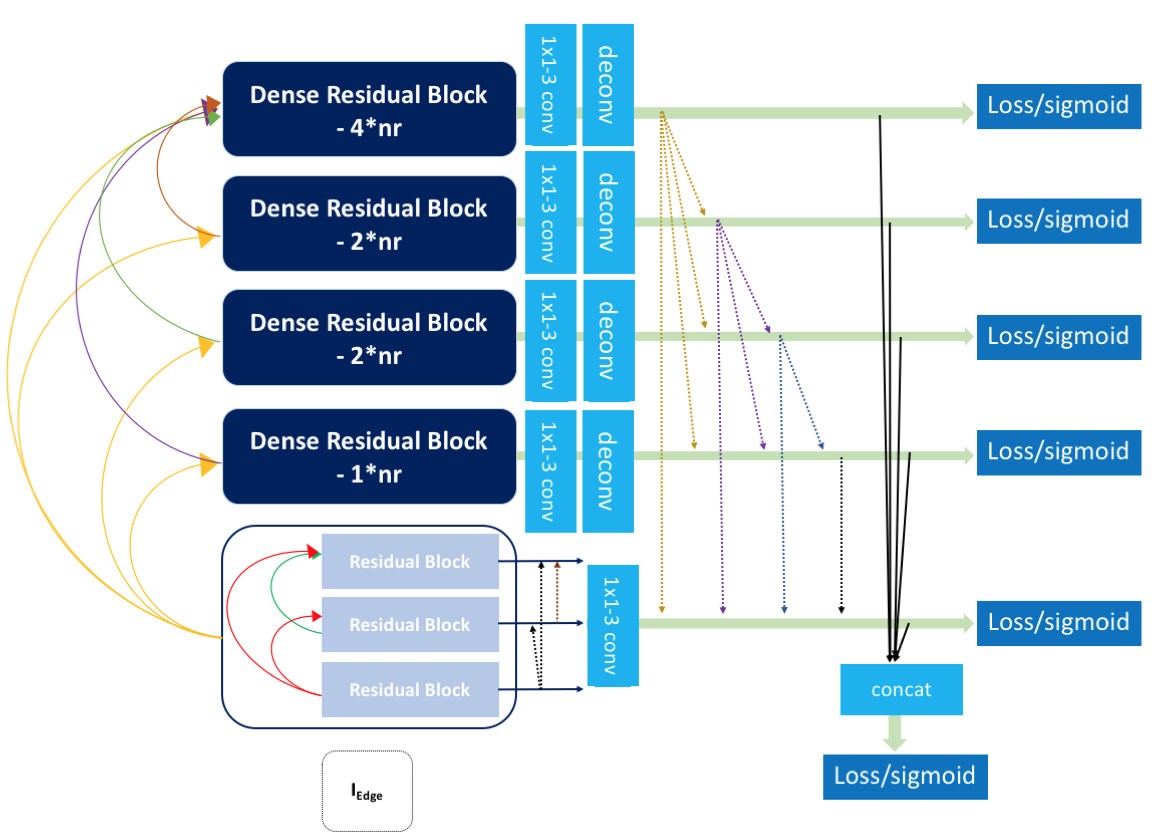}
	\caption{\protect Our proposed DenseEdgeNet architecture for edge detection.} 
	\label{fig:edgenet}
    \end{center}	
\end{figure*}

\subsection{DenseEdgeNet}

In order to train DenseEdgeNet, we generated images with ground truth for training by using Canny edge detector. Rather than taking $I_{SR}$ as an input, we took $I_{HR}$ as an input to train our network. We empirically found that the network that was trained with $I_{SR}$ as an input yielded poor performances possibly due to
wrong edge information contained in $I_{SR}$. 
 After training our proposed network, 
 $I_{Edge}$ were produced from $I_{SR}$.

\textbf{Network architecture.}
Inspired by previous edge detection method in deep learning~\cite{bertasius2015deepedge,liu2017richer,xie2015holistically}, we design our edge detection network by modifying RCF network~\cite{liu2017richer} as shown in Figure~\ref{fig:edgenet}. 
The RCF network, based on VGG16 network~\cite{simonyan2014very}, yielded state-of-the-art performance in edge detection. The RCF network consists of five stages, each of which generates a feature map through convolutional layers. We have replaced convolutional layers with dense residual blocks to increase the size of receptive field and to produce an enhanced feature map. Each stage of the residual block is designed to gradually increase the feature map channel size. We set \textit{nr} to 128 for the single complexity architecture in our experiments.
In addition to the skip connections in the block, we added a dense connection between the blocks to obtain a richer multi-scale feature map.

Inspired by HED~\cite{xie2015holistically}, we connected each dense residual block to the side output layer. This side output layer consists of one convolution layer with kernel size 1 and one upsampling layer. Before executing the convolution, we also added a dense connection instead of simply concatenating each side output. A deconvolutional layer was implemented by using pyramid pooling and pixel shuffle~\cite{lim2017enhanced} instead of conventional bilinear upsampling method. While only using bilinear upsampling for upscale can be difficult to utilize multi-scale feature map information, we could obtain more rich feature map information through pyramid pooling and pixel shuffle method.

\textbf{Short connection.}
Introducing dense connections in dense residual blocks enables us to increase the receptive field size. However, there is still a lack of global information regarding the side output. For these problems, we propose to % inspire us to
use both local and global information to predict edges. In~\cite{hou2017deeply}, introducing short connections between the side outputs resulted in generating more sophisticated saliency map. Motivated by this method, we also integrated short connection into each side output of dense residual blocks. With the short connection, lower side output that include local detail information can accurately predict edges and can refine edge maps using deeper side outputs. 

\textbf{Two independent branches.}
Similar to \cite{bertasius2015deepedge}, DenseEdgeNet consists of two independently trained branches. One branch learns to predict the likelihood of an edge by using binary cross entropy loss. This branch perform pixel-level edge classification by classifying whether a given point is edge or not. The other learns to predict the specific value of a given point by using $l1$ loss. The second branch performs a regressor to predict the exact values at a particular point. In~\cite{bertasius2015deepedge}, they demonstrated that utilizing both branches can produce better results when compared to using only one branch. To produce edge images, the output from two independent branches was averaged.

\textbf{Multi-complexity architecture.}
In previous deep-learning based edge detection methods~\cite{bertasius2015deepedge,liu2017richer}, they used multi-scale methods to obtain more accurate edges. We adopted this multi-complexity method to obtain rich details of an image. In the single complexity architecture, an initial feature map size, \textit{nr}, was set to 128. 

On the other hand, in the multi-complexity architecture, there are three independent models for three feature map sizes. \textit{eg}, \textit{$nr_1$}, \textit{$nr_2$},\textit{$nr_3$}. In the experiment, we set \textit{$nr_1$} = 32, \textit{$nr_2$} = 64 and \textit{$nr_3$} = 128 . Then, we trained independently each model and then merged together at the final classification. In the multi-complexity architecture , we could obtain a more accurate and clear edge by predicting edges from richer feature map information.

%-------------------------------------------------------------------------
\subsection{MergeNet}

\begin{figure}[b]
    \centering
	\includegraphics[width=1\linewidth]{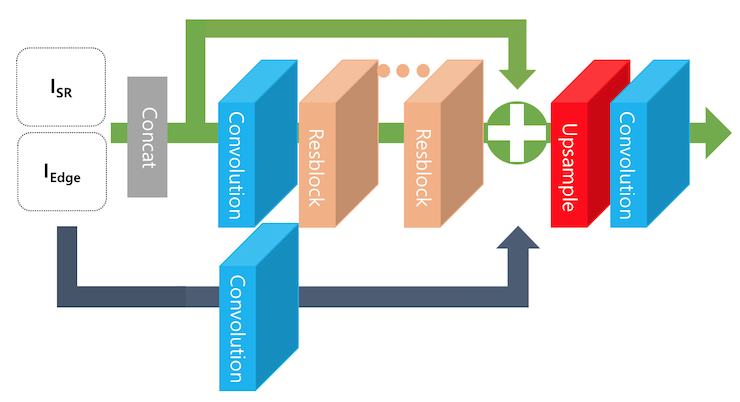}
	\caption{\protect Illustration of our proposed MergeNet. Unlike EDSR~\cite{lim2017enhanced} that uses only one skip connection, MergeNet have one more edge skip connection.} 
	\label{fig:mergenet}
\end{figure}

We propose MergeNet that integrates  $I_{SR}$ and $I_{Edge}$ and is based on one of state-the-art methods, EDSR~\cite{lim2017enhanced} for SR. The network architecture for MergeNet is shown in Figure~\ref{fig:mergenet}. EDSR takes an input image with 3 channels and uses residual learning. One the other hand, our MergeNet takes an input image with 4 channels (RGB + Edge) and add edge skip connection. An edge skip connection refers to connecting the edges of an input to residual learning. By increasing the feature map channel of edge map to the number of desired feature map channels through the convolution layer, we added it to the residuals together. This method enabled us to efficiently utilize the edge information for learning to integrate $I_{SR}$ and $I_{Edge}$. 

EDSR used 256 feature map channels and 32 residual blocks while MergeNet used 128 feature map channels and 16 residual blocks.  Although the number of parameters are reduced by half, the performance does not significantly differ from the output with original parameters and shows that the speed of training is faster. 

%-------------------------------------------------------------------------

\begin{table*}[t]
\begin{center}
\begin{tabular}{cccccccccc} 
\hline
          &       & \multicolumn{2}{c}{Set5}                              & \multicolumn{2}{c}{Set14}                             & \multicolumn{2}{c}{BSD100}                            & \multicolumn{2}{c}{Urban100}                          \\
Algorithm & scale & PSNR                      & SSIM                      & PSNR                      & SSIM                      & PSNR                      & SSIM                      & PSNR                      & SSIM                      \\ \hline

Bicubic   & 4     & 28.42                     & 0.810                     & 26.10                     & 0.704                     &25.96                    & 0.659                     & 23.15                     & 0.659                     \\
A+~\cite{timofte2014a+}        & 4     & 30.30                     & 0.859                     & 27.43                     & 0.752                     & 26.82                     & 0.710                     & 24.34                     & 0.720                     \\
SRCNN~\cite{dong2016image}     & 4     & 30.49                     & 0.862                     & 27.61                     & 0.754                     & 26.91                     & 0.712                     & 24.53                     & 0.724                          \\
FSRCNN~\cite{dong2016accelerating}    & 4     & 30.71                     & 0.865                     & 27.70                     & 0.756                     & 26.97                     & 0.714                     & 24.61                     & 0.727                          \\
VDSR~\cite{kim2016accurate}      & 4     & 31.35                     & 0.882                     & 28.03                     & 0.770                      & 27.29                     & 0.726                     & 25.18                     & 0.753                          \\
DRCN~\cite{kim2016deeply}      & 4     & 31.53                     & 0.884                     & 28.04                     & 0.770                    & 27.24                     & 0.724                     & 25.14                     & 0.752                          \\
DRRN~\cite{tai2017image}      & 4     & 31.68                     & 0.888                     & 28.21                     & 0.770                      & 27.38                     & 0.728                     & 25.44                     & 0.764                          \\
LapSRN~\cite{lai2017deep}    & 4     & 31.54                     & 0.885                     & 28.19                     & 0.772                      & 27.32                     & 0.728                     & 25.87                     & 0.756                           \\
EDSR~\cite{lim2017enhanced}      & 4     & {\bf  32.46     }                & {\bf 0.897  }                   & {\bf 28.80     }                & {\bf 0.788          }            & {\bf 27.61}                     & 0.737                     & {\bf  26.25 }                    & {\bf 0.795   }                       \\
D-DBPN ~\cite{haris2018deep}   & 4     & {\bf 32.47    }                 & {\bf 0.898      }               & {\bf 28.82     }                & {\bf 0.786    }                 & {\bf 27.72 }                    & {\bf 0.740  }                   & {\bf 27.08 }                    & {\bf 0.795    }                 \\\hline
Ours      & 4     & 31.02                     & 0.892                     & 27.24                     & 0.780                       &      27.06             & {\bf 0.738   }                          & 25.82                     & 0.791                           \\\hline
Bicubic   & 8     & 24.39                     & 0.657                     & 23.19                     & 0.568                     & 23.67                     & 0.547                     & 20.74                     & 0.516                     \\
A+~\cite{timofte2014a+}        & 8     & 25.52                     & 0.692                     & 23.98                     & 0.597                     & 24.20                     & 0.568                     & 21.37                     & 0.545                     \\
SRCNN~\cite{dong2016image}     & 8     & 25.33                     & 0.689                     & 23.85                     & 0.593                     & 24.13                     & 0.565                     & 21.29                     & 0.543                     \\
FSRCNN~\cite{dong2016accelerating}    & 8     & 25.41                     & 0.682                     & 23.93                     & 0.592                     & 24.21                     & 0.567                     & 21.32                     & 0.537                     \\
VDSR~\cite{kim2016accurate}      & 8     & 25.72                     & 0.711                     & 24.21                     & 0.609                     & 24.37                     & 0.576                     & 21.54                     & 0.560                     \\
LapSRN~\cite{lai2017deep}    & 8     & 26.14                     & 0.738                     & 24.44                     & 0.623                     & 24.54                     & 0.586                     & 21.81                     & 0.582                     \\
EDSR~\cite{lim2017enhanced}      & 8     & 26.97                     & 0.775                     & {\bf 24.91 }                    & 0.640                     & 24.80                     & 0.596                     & 22.47                     & 0.620                     \\
D-DBPN ~\cite{haris2018deep}    & 8     & \multicolumn{1}{l}{{\bf 27.21}} & \multicolumn{1}{l}{{\bf 0.784}} & \multicolumn{1}{l}{{\bf 25.13}} & \multicolumn{1}{l}{{\bf 0.648}} & \multicolumn{1}{l}{{\bf 24.88}} & \multicolumn{1}{l}{{\bf 0.601}} & \multicolumn{1}{l}{{\bf 23.25}} & \multicolumn{1}{l}{{\bf 0.622}} \\\hline
Ours      & 8     & \multicolumn{1}{l}{{\bf 27.12}} & \multicolumn{1}{l}{{\bf 0.786}} & \multicolumn{1}{l}{24.81} & \multicolumn{1}{l}{{\bf 0.643}} & \multicolumn{1}{l}{{\bf 24.84}} & \multicolumn{1}{l}{{\bf 0.602}} & \multicolumn{1}{l}{{\bf 22.56}} & \multicolumn{1}{l}{{\bf 0.626}} \\
\hline
\end{tabular}
\end{center}
\caption{\protect Quantitative evaluation of SR algorithms. Average PSNR/SSIM for scale factors $\times$4 and $\times$8. {\bf Bold} indicates the first and the second best performances.}
\label{table2}
\end{table*}

\section{Experimental Results}
\subsection{Dataset}

We use the DIV2K dataset~\cite{timofte2017ntire} to train all of our models. DIV2K is a high quality (2K resolution) image data set from CVPR NTIRE 2017 challenge. DIV2K dataset consists of 800 training images, 100 validation images, and 100 test images. We train our models with 800 training images and only use 10 validation images (801 to 810) in training process. For evaluation, we use four standard benchmark datasets: Set5~\cite{bevilacqua2012low}, Set14~\cite{zeyde2010single}, BSD100~\cite{martin2001database}, Urban100~\cite{huang2015single}. Each dataset has different properties. Set5, Set14, and BSD100 are composed of natural scenes. Urban100 dataset includes urban scenes and has many vertical and horizontal lines.  SR results were evaluated with PSNR and SSIM~\cite{wang2004ssim} on Y channel of transformed YCbCr space.

\subsection{Implementation and training details}
\textbf{Offset problem.}
When upsampling $\times$8 downsampled images of the DIV2K dataset~\cite{timofte2017ntire}, the size of SR images did not exactly match the size of ground truth images. Although it is a small difference, it seems to cause significant offset problems. That's because it is very sensitive to incorporating edge information with SR images. To prevent this issue, we resized the ground truth images whose width and height are multiples of 8, and then downsampled it using bicubic interpolation.

\textbf{Training settings.}
In each training batch, we randomly extracted 16 $I_{LR}$ patches with the size of 48 $\times$ 48 as inputs. Data augmentation was performed on 800 training images, which are randomly rotated by 90$^\circ$ and flipped horizontally or vertically. We implemented our proposed networks with PyTorch framework and trained them using ADAM optimizer~\cite{kingma2014adam}. An initial learning rate was set to $10^{-4}$ for both SRNetwork and MergeNet. Then, we decrease them by half for every 100 epochs. For DenseEdgeNet, a learning rate is initialized to $10^{-6}$ and fix it during the whole training process. Training our proposed networks roughly takes 2 days with a NVIDIA Titan Xp GPU for 200 epochs.

\subsection{Analyses on models}

\textbf{Speed.}
When compared to other state-of-the-art networks, the speed of our network is comparable. D-DBPN\cite{haris2018deep} takes about 50 seconds per image, and EDSR\cite{lim2017enhanced} takes about 8.3 seconds per image. %On the other hand, t
The speed of our networks correspond to about 23 seconds per image. 
Because our network produces images in sequential steps, the speed of our network is slower %take longer 
than EDSR, but much faster than %less time than 
D-DBPN as summarized in Table~\ref{table3}.

%-------------------------------------------------------------------------
\begin{table}[!h]
\begin{center}
\begin{tabular}{|l|l|l|}
\hline
Dataset                 & Algorithms & Speed                        \\ \hline
                        & EDSR       & {\color[HTML]{FE0000} 8.5s}  \\ \cline{2-3} 
                        & D-DBPN      & 50s                          \\ \cline{2-3} 
\multirow{-3}{*}{DIV2K} & Ours       & {\color[HTML]{3531FF} 22.4s} \\ \hline
\end{tabular}
\end{center}
\caption{\protect Comparisons of state-of-the-art methods on speed with DIV2K dataset. {\color[HTML]{FE0000}Red} indicates the best results and the {\color[HTML]{3531FF}blue} indicates the second best results.}
\label{table3}
\end{table}
%-------------------------------------------------------------------------

\begin{figure}[!h]
    \begin{center}
    \centering
	\includegraphics[width=0.8\linewidth]{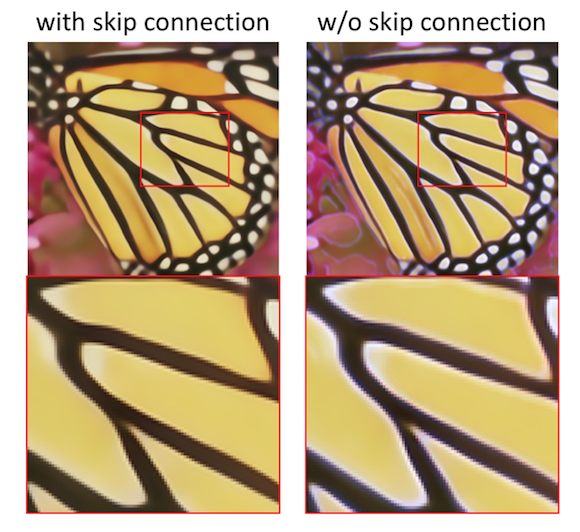}
	\caption{\protect Visual comparisons with, without edge skip connection.} 
	\label{fig:skip}
    \end{center}	
\end{figure}

\textbf{Effectiveness of edge skip connection.}
We introduce edge skip connection to reinforce the edge information in MergeNet. This method improved the performance of the networks as shown in Figure~\ref{fig:skip}. Without edge skip connection, the edge is clear and sharp, but cannot merged the edge with the interpolated image. Thus, images with overall color differences and artifacts were yielded. 
In contrast, the visual results with edge skip connection were qualitatively improved while maintaining the sharpness of edges. 
This shows that although edge information helps to reconstruct high frequency information, it is important  how to well incorporate both interpolated image and edge map. 

\begin{figure}[!h]
    \begin{center}
    \centering
	\includegraphics[width=1\linewidth]{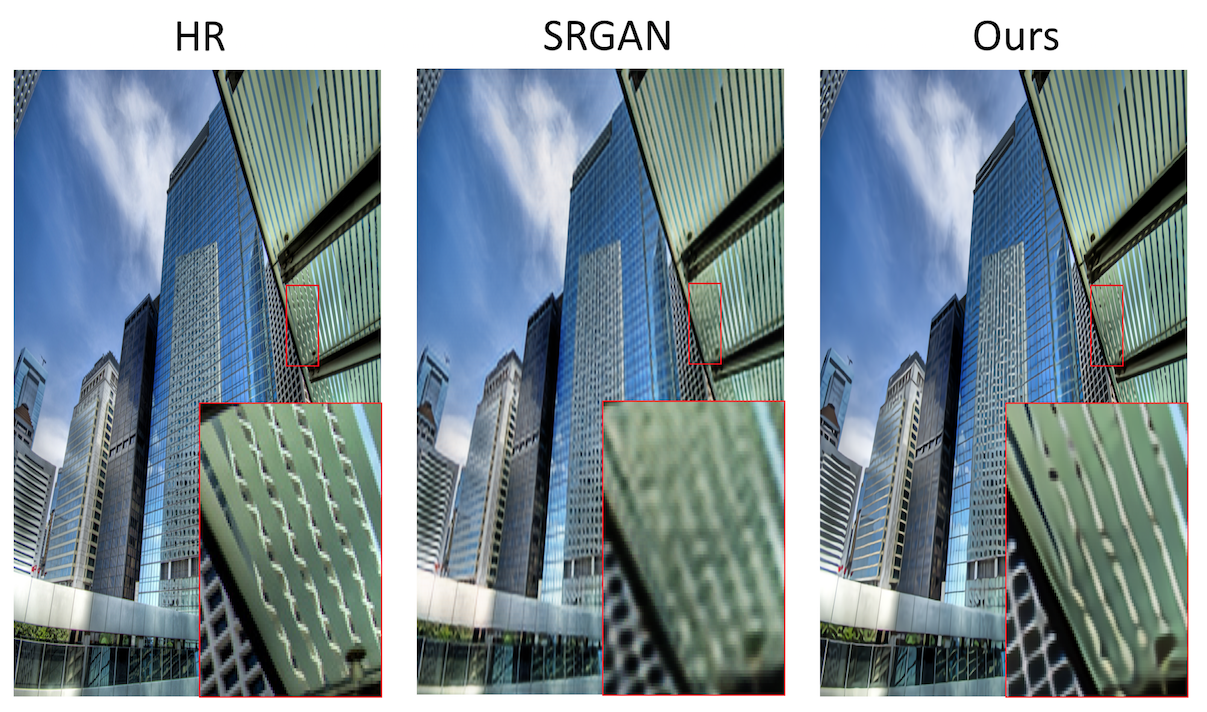}
	\caption{\protect Results of SRGAN, our SREdgeNet for $\times$8 enlargement.}
	\label{fig:srgan}
    \end{center}
\end{figure}

\textbf{Comparison with SRGAN.}
We compared our SREdgeNet with SRGAN~\cite{ledig2017photo} qualitatively and quantitatively.
Figure~\ref{fig:srgan} shows the SR results for image img\_061 from Urban100. 
SRGAN reconstructed the overall pattern well, but the detailed edges were not sharp enough. On the other hand, our proposed SREdgeNet network was able to recover sharp edges and overall patterns.
Table~\ref{tbl_srgan} reports the quantitative comparisons between SRGAN and our SREdgeNet. 
For $\times$4 enlargement, our SREdgeNet network gains are  0.77 dB and 1.92 dB higher than SRGAN on the BSD100 and Urban100, respectively. On $\times$8 enlargement, 
the results were also better than SRGAN. Our network has 0.82 dB and 1.32 dB higher PSNR than SRGAN on the BSD100 and Urban100, respectively.

%-------------------------------------------------------------------------
\begin{table}[!h]
\begin{center}
\begin{tabular}{cccccc}
\hline
          &       & \multicolumn{2}{c}{BSD100}                                  & \multicolumn{2}{c}{Urban100}                                \\
Algorithm & Scale & PSNR                         & SSIM                         & PSNR                         & SSIM                         \\\hline
SRGAN     & 4     & 26.29                        & 0.698                        & 23.90                        & 0.700                        \\
Ours      & 4     & {\color[HTML]{FE0000} 27.06} & {\color[HTML]{FE0000} 0.738} & {\color[HTML]{FE0000} 25.82} & {\color[HTML]{FE0000} 0.791} \\\hline
SRGAN     & 8     & 24.02                        & 0.561                        & 21.24                        & 0.543                        \\
Ours      & 8     & {\color[HTML]{FE0000} 24.84} & {\color[HTML]{FE0000} 0.602} & {\color[HTML]{FE0000} 22.56} & {\color[HTML]{FE0000} 0.626}
\\\hline
\end{tabular}
\end{center}
\caption{\protect Comparison of the SRGAN and Ours on $\times$4 and $\times$8 enlargement. {\color[HTML]{FE0000}Red} indicates the best results.}
\label{tbl_srgan}
\end{table}
%-------------------------------------------------------------------------

\begin{figure*}[t]
    \begin{center}
    \centering
	\includegraphics[width=0.85\linewidth]{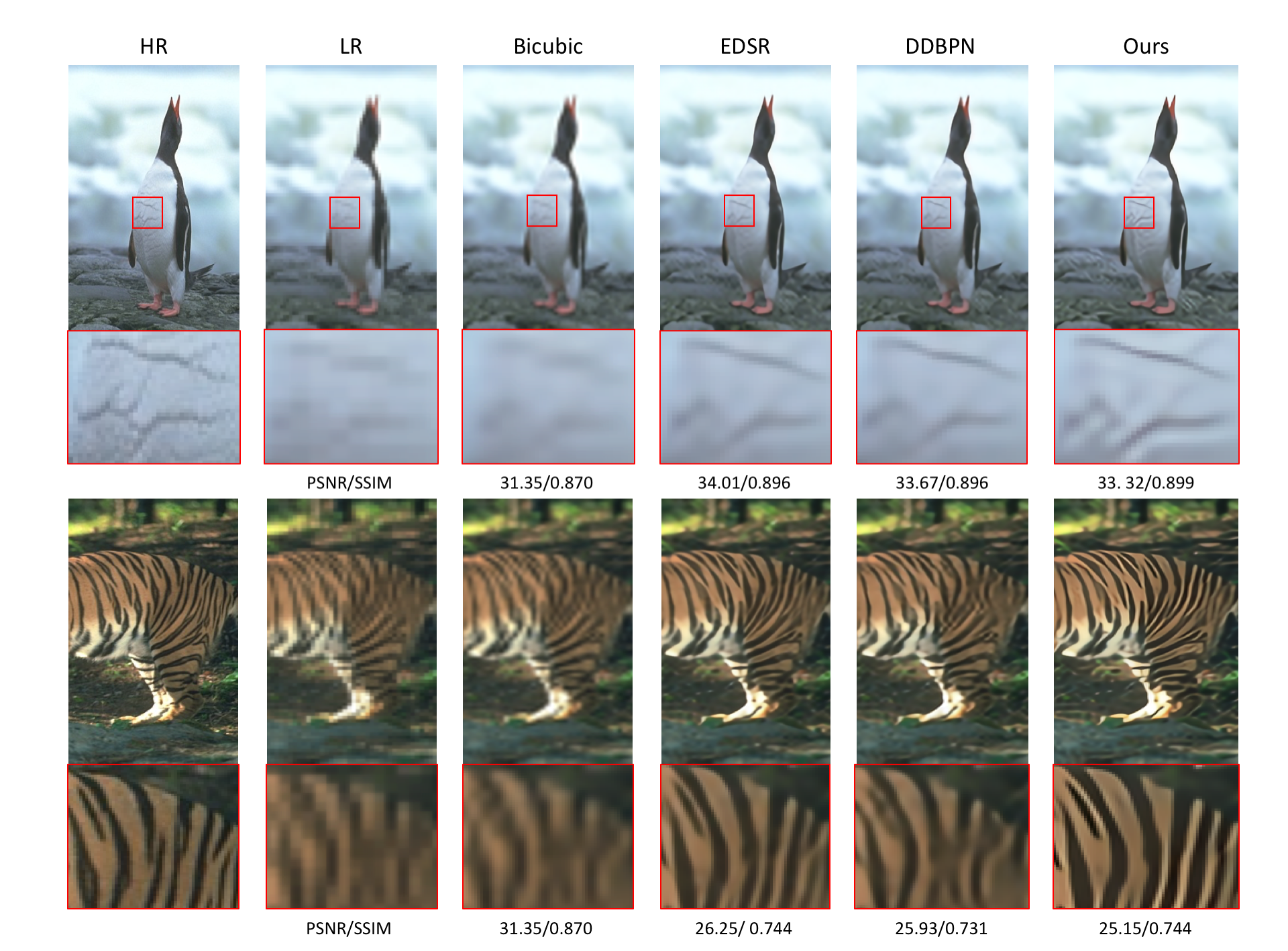}
	\caption{\protect Qualitative comparison of our models with other works on $\times$4 super-resolution. The SR results are for 106024 and 108005 from BSD100, respectively.}
	\label{fig:comparex4}
    \end{center}	
\end{figure*}
\begin{figure*}[t]
    \begin{center}
    \centering
	\includegraphics[width=0.8\linewidth]{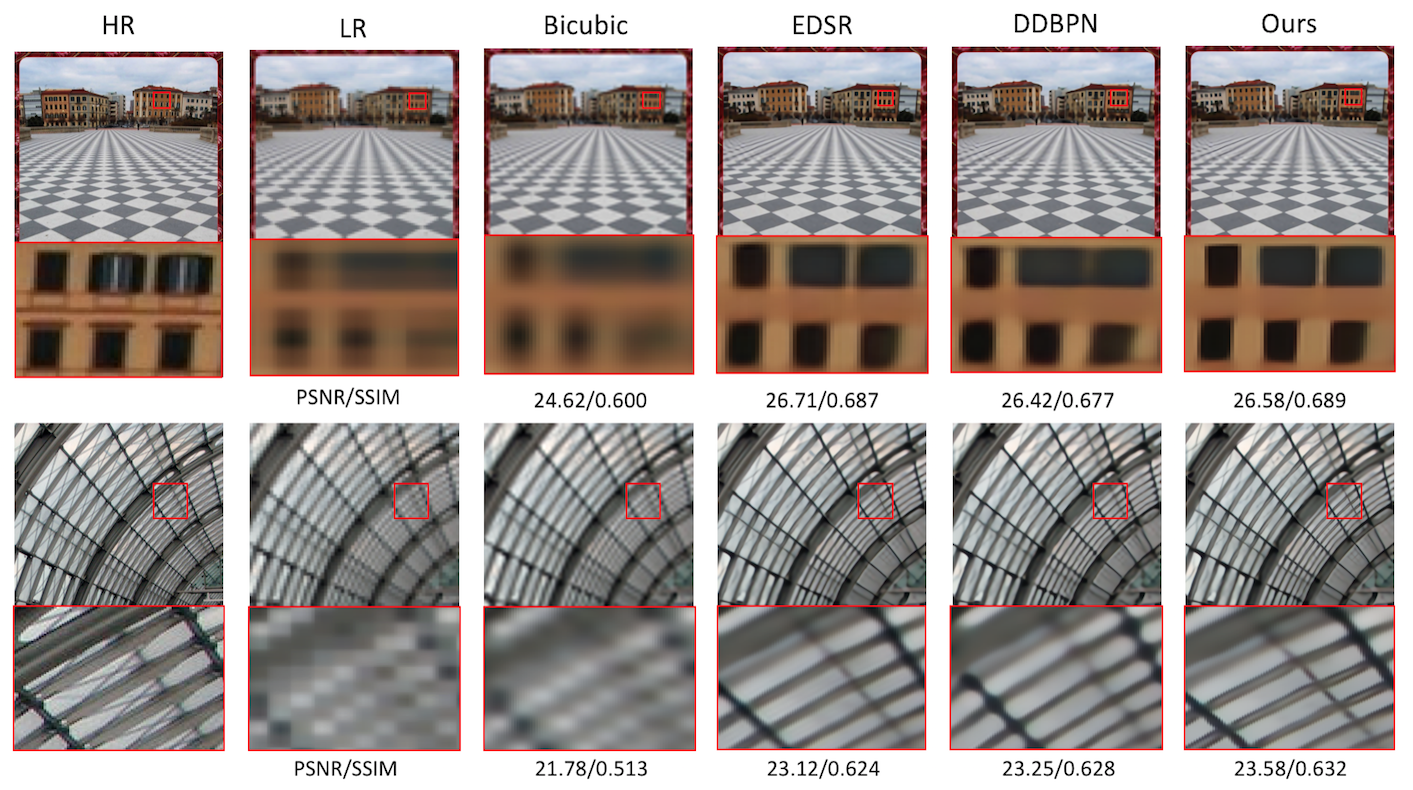}
	\caption{\protect Qualitative comparison of our models with other works on $\times$8 SR. SR results are for image img\_021 and img\_002 from Urban100, respectively.}
	\label{fig:comparex8}
    \end{center}	
\end{figure*}

\subsection{SR results on benchmark datasets}

We compared our SREdgeNet with nine state-of-the-art networks: A+~\cite{timofte2014a+}, SRCNN~\cite{dong2016image}, FSRCNN~\cite{dong2016accelerating}, VDSR~\cite{kim2016accurate}, DRCN~\cite{kim2016deeply}, DRRN~\cite{tai2017image}, LapSRN~\cite{lai2017deep}, EDSR~\cite{lim2017enhanced}, D-DBPN~\cite{haris2018deep}. 
Note that the goal of our proposed SREdgeNet is to improve qualitative SR image quality while maintaining quantitative results in terms of PSNR or SSIM.
In this sense, the quantitative results of our SREdgeNet is encouraging as shown in Table~\ref{table2}. 
Our proposed SREdgeNet yielded comparable SSIM to other state-of-the-art methods such as EDSR or D-DBPN within 0.008 difference max for $\times$4 enlargement.
For $\times$8 enlargement, our proposed method achieved the first or the second best performances over other state-of-the-art methods in both PSNR and SSIM.
In particular, for the Urban100 dataset~\cite{huang2015single} that consists of many repetitive patterns and edges are prominent, our network has advantage for utilizing edge information over other methods.
Our proposed method yielded the third best SSIM among all comparing methods (0.791) that was comparable to the best performance (0.795) for $\times$4 enlargement, while
achieved the best SSIM (0.626) for $\times$8 enlargement. 
%for the Urban100~\cite{huang2015single}, our results are similar to LapSRN~\cite{lai2017deep} in PSNR, but our proposed network obtains 0.791 which is 0.035 better than LapSRN. Comparing to EDSR~\cite{lim2017enhanced}, EDSR has only 0.004 higher than our proposed network. 

The visual results of $\times$4 enlargement are shown in Figure~\ref{fig:comparex4}. Qualitatively, we observed that other networks yielded noticeable artifacts and blurred edges to reconstruct particular patterns. In contrast, our proposed SREdgeNet network can recover sharper and clear edges, resulting in much better visual quality. It shows that edges obtained from DenseEdgeNet seem to guide for recovering HR components.

Our network shows its effectiveness in scaling factor $\times$8. Especially, our network achieved the best SSIM results in most test datasets.
% all fields except one dataset. 
The most prominent results are also shown on Urban100 dataset for $\times$8 enlargement. The PSNR of our network is about 0.7 dB lower than D-DBPN~\cite{haris2018deep}, but the SSIM is higher than D-DBPN. Compared to $\times$4 enlargement, the PSNR difference with other networks was reduced. It shows that the larger the upscaling factor is, the more effective our network is.

The visual results of $\times$8 enlargement are shown in Figure~\ref{fig:comparex8}. The results of D-DBPN show that the boundaries between windows are blurred and cannot be distinguished. EDSR produces more qualitatively good results but still yields blurred images. In contrast, our network produced sharper and clear edges and reconstructed regular patterns well. We found that since the HR component of the $\times$8 downsampled image is deficient, it seems helpful to merge the edges containing the HR information with interpolated image.

\section{Conclusion}

In this paper, we proposed SREdgeNet, an edge enhanced single image SR network that consists of three modules: SR module, edge detection module, and merge module. The first module, EDSR*, a variant of ESDR incorporated pyramid pooling into ESDR, yielded better performance than EDSR.

We also proposed DenseEdgeNet, the second module, which is composed of dense residual blocks and has a dense connection to get richer feature information. We also introduced a multi-complexity method into DenseEdgeNet to yield better edge images. DenseEdgeNet yields edges from the output of the first SR module. 

Thirdly, we proposed MergeNet that integrates the outputs of the first and second modules to yield edge enhanced SR images. This network is based on EDSR and we added an edge skip connection so that leads to well-fused upscaled images and edges.

Finally, our proposed method qualitatively outperformed against previous state-of-the-art methods. SREdgeNet shows state-of-the-art performances in terms of SSIM while maintaining comparable PSNR for $\times$8 enlargement with reasonably fast computation time.

\section*{Acknowledgments}

This work was supported partly by 
Basic Science Research Program through the National Research Foundation of Korea (NRF) 
funded by the Ministry of Education (NRF-2017R1D1A1B05035810), partly by 
the Technology Innovation Program or Industrial Strategic Technology Development Program 
(10077533, Development of robotic manipulation algorithm for grasping/assembling 
with the machine learning using visual and tactile sensing information) 
funded by the Ministry of Trade, Industry \& Energy (MOTIE, Korea), and partly by 
a grant of the Korea Health Technology R\&D Project through the Korea Health
Industry Development Institute (KHIDI), funded by the Ministry
of Health \& Welfare, Republic of Korea (grant number : HI18C0316).

{\small
\bibliographystyle{ieee}
\bibliography{egbib}
}

\end{document}